# Dense–SwinV2: Channel-Attentive Dual-Branch CNN–Transformer Learning for Cassava Leaf Disease Classification

Shah Saood[1] , Saddam Hussain Khan[2*]

[1]Artificial Intelligence Lab, Department of Computer Systems Engineering, University of Engineering and Applied Sciences (UEAS), Swat 19060, Pakistan

[2]Interdisciplinary Research Center for Smart Mobility and Logistics (IRC-SML), King Fahd University of Petroleum and Minerals (KFUPM), Dhahran 31261, Saudi Arabia.

Email: saddam.khan@kfupm.edu.sa

## Abstract

Cassava leaf disease is difficult to detect, since even expert agricultural visual inspections are error-prone. The proposed approaches for diagnosis are very time-consuming and require expert knowledge; therefore, there exists a need for the automation of these tasks. In this regard, this work presents a new Hybrid Dense–SwinV2, a two-branch framework that jointly leverages densely connected convolutional features and hierarchical customized Swin Transformer V2 (SwinV2) representations for cassava disease classification. The proposed framework captures high-resolution local features through its DenseNet branch, preserving the fine structural cues and also allowing for effective gradient flow. Concurrently, the customized SwinV2 models global contextual dependencies through the idea of shifted-window self-attention, which enables the capture of long-range interactions critical in distinguishing between visually similar lesions. Moreover, an attention channel-squeeze module is employed for each CNN-Transformer stream independently to emphasize discriminative disease-related responses and suppress redundant or background-driven activations. Finally, these discriminative channels are fused to achieve refined representations from the Dense local and SwinV2 global correlated strengthened feature map, respectively. The proposed Dense-SwinV2 utilized a public cassava leaf disease dataset of 31,000 images, comprised of five diseases, including Brown Streak, Mosaic, Green Mottle, Bacterial Blight, and normal leaf conditions. The proposed Dense-SwinV2 demonstrates a significant classification accuracy of **98.02%** with an F1-score of **97.81%**, outperforming well-established convolutional and transformer models. These results underline the fact that Hybrid Dense–SwinV2 offers robustness and practicality in the field-level diagnosis of cassava disease, and real-world challenges related to occlusion, noise, and complex backgrounds.

## 1. Introduction

Cassava is among the staple crops that are very important in steamy and semitropical areas, particularly in sub-Saharan Africa, as it provides a food security safety net and supports rural livelihoods[1]. Cassava resists poor soils and drought conditions; hence, it remains a vital crop for over ten million smallholder farmers growing it for both food and cash [1]. Worldwide production stands at over 324 million tonnes, with the African continent contributing the largest share [2]. Despite this hardy nature against bad weather conditions, its cultivation faces a serious menace from viral and bacterial pathogens [2], [3]. The most devastating diseases include Cassava Mosaic Disease (CMD) and Cassava Brown Streak Disease (CBSD), diseases that reduce crops by more than 35% and result in combined annual losses of over one billion USD in the affected economies [4], [5]. This calls for timely and efficient disease diagnosis to assure a sustainable level of productivity, hence livelihood [6]. Accurate identification of cassava diseases within precision crop management is an important factor impacting the efficiency of crop production and food security. Traditional diagnosis for many diseases relies on field physical inspections or manual analysis by experienced experts or farmers, which is always time-consuming, costly, and sometimes inapplicable in resource-poor areas. Moreover, manual analysis is inefficient, especially for large covering regions, and not practical for farmers with minimal means and regions that can be considered remote in terms of geographical location [7].

As a countermeasure, research and development on precision agriculture systems have begun to integrate the use of unmanned aerial vehicles (UAVs) technology for crop surveillance over a large area. The UAV technology equipped with very efficient imaging technology, like RGB cameras, multi-spectral cameras, and thermal imaging cameras, can detect initial symptoms of diseases beyond human observation and implement treatments [8], [9]. However, imaging data captured through the usage of UAV technology also poses new challenges for processing the images for automated analysis purposes. Therefore, the agricultural vision community has increasingly opted for deep learning (DL) as one of the most promising alternatives for the automation of plant disease identification[10], [11], [12].

DL has played an important role in diverse fields, including automatic agriculture production and disease detection, medical diagnosis, NLP, security, commerce, etc [13], [14]. Deep convolutional neural networks (CNNs) are a subtype of DL CNNs that capture automatically hierarchical local features and are effectively employed in agriculture, especially for plant and leaf disease diagnosis[15]. Deep CNNs identify patterns on several visual levels, such as the texture and pigments on plant leaves, up to distinctive lesion formations. Moreover, Vision Transformer (ViT) has the ability to learn long-range contextual dependencies and achieved optimal performance[16], [17]. The Transformer is suitable for high-resolution UAV imagery due to its hierarchical design for global contextual feature extraction and integration with shifted-window attention [18].

However, the standard transformer various varients models require a stack of computational resources and significant volumes of data for training, rendering deployment on mobile or edge devices impractical. Moreover, several techniques have been developed; however, they lack robustness pertaining to real-world complications, like face occlusions of leaves, complex backgrounds, morphological variability, and issues with computational load. All of these modules are parallelized to achieve enhanced discriminability of visually similar disease phenotypes while maintaining the operational feasibility for their deployment under field environments[1], [19], [20]. Figure 1 represents a detailed diagram of the entire framework of the proposed method of cassava leaf disease diagnosis, which includes a detailed description of how an input image is processed through a series of steps to arrive at a final classification of a disease out of a total of five types of diseases affecting cassava leaves. As can be seen in Figure 1, input images of cassava leaves are first subjected to data augmentation to increase their generalization capability, after which features are extracted through parallel DenseNet and Swin Transformer V2 (SwinV2) branches.

This paper presents a hybrid architecture of a dual-branch comprising customized DenseNet and (SwinV2) blocks for hierarchical local feature abstraction and global contextual details, respectively. Moreover, Feature maps of individual CNN and Transformer models are refined through the use of a Multi-Scale squeezing and Attention block to highlight disease-relevant features and suppress redundant or background information. Finally, the Feature Fusion Module merges both models' discriminative channels, thereby allowing for the interactional enhancement between local and global cues. The contributions of this paper are:

- A new Dual-Branch Hybrid DenseNet-SwinV2 is introduced for fine-grained local textures and long-range contextual dependencies learning towards robust cassava disease classification. The customized SwinV2 block global interactions through hierarchical shifted-window self-attention, strengthening discriminative disease representation. Moreover, DenseNet with residual block skip connection reduces the vanishing gradient effectively and improves the convergence of the framework.
- A lightweight Multi-Scale and channel-squeezing attention mechanism is embedded within each model to enhance the salient disease patterns and suppress the background noise. The discriminative channels of each attention-refined Dense and SwinV2 maps are fused to achieve the refined diverse feature representations and preserve complementary local-global interactions[21].
- The proposed Dense-SwinV2 technique has shown contemporary performance on the heterogeneous clutter and occluded cassava disease dataset, with a significant improvement over well-established CNN and Transformer models. The technique is robust in real agricultural environments and a reliable framework for automatic in-field diagnosis of cassava diseases under complex environmental conditions[22], [23].

The related work is introduced in Section 2; The related work is introduced in Section 2; the proposed cassava disease classification framework is explained in Section 3.

## 2. Related Work

The traditional process of disease identification is carried out by actual physical observations or manual analysis of the expert or farmers. However, manual analysis is still widely used today and requires a significant amount of effort and time [1]. Therefore, DL based automatic high-resolution UAV imagery analysis has been employed in the previous studies. Pre-trained CNN architectures such as VGG19, InceptionResNetV2, DenseNet, and EfficientNet have achieved high validation accuracies for cassava disease classification, often above 85% [24]. It should be noted that despite the advantages of CNNs, they are vulnerable to overfitting. Overfitting can lead to a significant loss in performance accuracy if the conditions in the field deviate from those in the training set. One more disadvantage of CNNs is that they are not capable of extracting long-range spatial relationships. This capability is essential for differentiating diseases that have symptoms that are not only distant but also overlapping.

Recently, researchers have also explored the possibilities of using a combination of CNNs and Transformers. Such models have also been successful in achieving classification accuracy rates of more than 95%. However, some of the models require significant computational power and cannot be deployed in the field. Models like MobileNetV2 have some advantages [25]. The literature gap here is that there has not yet been any model that has the guarantee of the following: strong accuracy in the field conditions (in the presence of noise, occlusion, and complex backgrounds), computational efficiency in the edge environment, and the ability to combine local fine-grained features with contextual information [26]. Despite the guarantee of strong local feature extraction of CNNs and the strong global perspective of Swin Transformers (SwinT), the parametric efficiency and the synergistic combination of these two models have not yet been thoroughly investigated in the context of cassava disease detection [20], [27].

This work will bridge this gap by proposing a new hybrid and dual-branch network architecture that can incorporate DenseNet to utilize the dense local features and SwinV2 to utilize global features. This work will also contribute a new module called Multi-Scale Attention to refine the features of both branches, and another module called Feature Fusion Module to synergistically utilize local and global features. It balances accuracy, robustness, and computational efficiency; hence, this model presents a practical solution for an automated cassava disease diagnosis system under real-world field conditions [3], [28]. Table 1 below shows a comprehensive summary of existing deep learning techniques in cassava disease classification by comparing eight different studies from existing literature. It is clear from Table 1 that existing techniques, such as VGG19 with a maximum accuracy of 80.27%, InceptionResNetV2 with a maximum accuracy of 87.00%, and other techniques like CNN-CA with a maximum accuracy of 75.00% in real-world challenges, are limited in performance due to overfitting and underfitting in training data. More recent techniques, such as MAIANet with a maximum accuracy of 95.83% and Attention-EfficientNet with a maximum accuracy of 96.30%, are limited in real-world challenges due to high computational cost and scalability issues, in real-world challenges and also in integrating fine-grained local features with global context information, which is addressed in our proposed model, Hybrid Dense-SwinV2.

Table 1: Existing study summary.

| Ref. | Model | Acc. (%) | Dataset | Key Limitation(s) |
|---|---|---|---|---|
| Alford & Tuba (2024) [11] | TL-VGG19 | 80.27% | Cassava leaf image dataset | Overfitting limits generalization. |
| Singh et al. (2023) [12] | InceptionResNetV2 | 87.00% | Cassava disease dataset | Insufficient training. |
| Tewari & Kumari (2024) [5] | Lightweight CNN-CA | 75.00% | Field images of cassava leaves | Struggled with a real challenge. |
| Zhang et al. (2024) [13] | MAIANet (Multi-Attention) | 95.83% | Cassava leaf dataset | High computational cost, unsuitable for low-supply deployment. |
| Ravi et al. (2021) [14] | Attention- EfficientNet | 96.30% | Cassava dataset | Scalability problems due to incomplete dataset diversity. |
| Sholihin et al. (2023) [15] | AlexNet - SVM | 90.70% | Cassava leaf images | Dataset size affects scalability. |
| Sapre et al. (2023) [20] | ANN | Suboptimal | Imbalanced cassava dataset | Poor generalization ability. |

## 3. Methodology

To make use of convolutional and transformer features for effective cassava disease classification, we develop a research work incorporating a hybrid Dense-SwinV2 model. Figure 2 is a complete architectural blueprint of the proposed Hybrid Dense-SwinV2 architecture, with a great level of detail being presented for the dual-branch architecture. Three distinct vertical regions are identified in the figure: one for the DenseNet branch on the left for local feature extraction, one for the SwinV2 branch on the right for global feature extraction, and one for the Multi-Scale Attention and Feature Fusion module in the middle. The DenseNet branch makes use of dense connectivity, wherein the feature maps from all the preceding convolutional layers are used as input to the current layer, allowing for effective gradient flow as well as feature reuse. The SwinV2 branch makes use of hierarchical stages with shifted window attention, allowing the model to capture the long-range spatial dependencies in the images, which are very useful in discriminating between very similar disease patterns. The outputs of the DenseNet branch as well as the SwinV2 branch are fed into the Multi-Scale Attention Blocks, which help in refining the disease-relevant features while suppressing the background noises, followed by the Feature Fusion Module, which combines the features synergistically. To overcome some limitations of the dataset and make the system more robust, we also implement data augmentation during preprocessing. The overall architecture can be broken down into two parallel processing paths, one for feature extraction and another for feature enhancement [29]. In the DenseNet processing path, it focuses on the detailed feature extraction based on high-resolution leaf images, and in the SwinV2 processing path, it learns the global context information from the images using a shifted window self-attention mechanism, which is important for distinguishing leaf images that have a similar symptomatic presentation. Feature maps produced by every channel are passed through a special attention mechanism and channel squeeze called Multi-Scale Attention and Channel-Squeeze while undergoing refinement before being fused. These features improve disease-related activations while discarding unnecessary information about irrelevant background detail [30]. In our assessment procedure, the performance comparison is conducted based on three types of models: the proposed Hybrid Dense-SwinV2 model, the current best vision transformer architectures, and the recent CNN architectures. A graphical version of the entire procedure is explained in Figures 1 and 2.

### 3.1. Data Augmentation

Data augmentation is one of the most important ways to increase model generalization, especially when data is limited. Exposing machine and deep learning models to synthetically augmented datasets makes them more robust and less prone to overfitting. Empirical studies on image-based classification tasks always attest that strategic augmentation leads to reduced error rates. In practice, augmentation emulates the natural variability of the real world, enhancing the capability of a model to identify a wide range of samples.[31] Thus, the augmentation technique is put into practice in the cassava leaf dataset; there is a high class imbalance, especially concerning the Cassava Mosaic Disease class. Doing this could result in a biased model, increased false alarms, and deterioration in general performance. Augmentation reduces this problem by artificially increasing the minority classes. It allows random horizontal and vertical flips, scaling, reflection, and shearing techniques. These transformations increase the diversity of the dataset without altering the pathological semantics and hence force the network to learn invariant features. Table 2 gives a detailed statistical description of the Cassava leaf disease dataset used in this research. The Cassava leaf disease dataset contains 31,179 high-resolution images, distributed across five class labels or disease types. The class labels are Cassava Bacterial Blight, containing 7,322 images;

Cassava Brown Streak, containing 6,695 images; Cassava Green Mottle, containing 7,018 images; Healthy leaves, containing 6,330 images; and Cassava Mosaic Disease, containing 3,814 images. Furthermore, the table shows the split for the train and test sets as 81%-19%, with 25,441 images for the train set and 5,738 for the test set. Additionally, the table shows the standard input size as 224x224x3. This comprehensive documentation reveals important class imbalance characteristics, particularly the underrepresentation of Cassava Mosaic Disease (3,814 samples) compared to other classes, which justifies our data augmentation strategy to artificially expand minority classes and prevent biased model development. The dataset's substantial size and realistic class distribution, combined with natural variations in occlusion, background clutter, and lighting conditions, provide a challenging and representative benchmark for evaluating cassava disease classification under real-world field conditions.

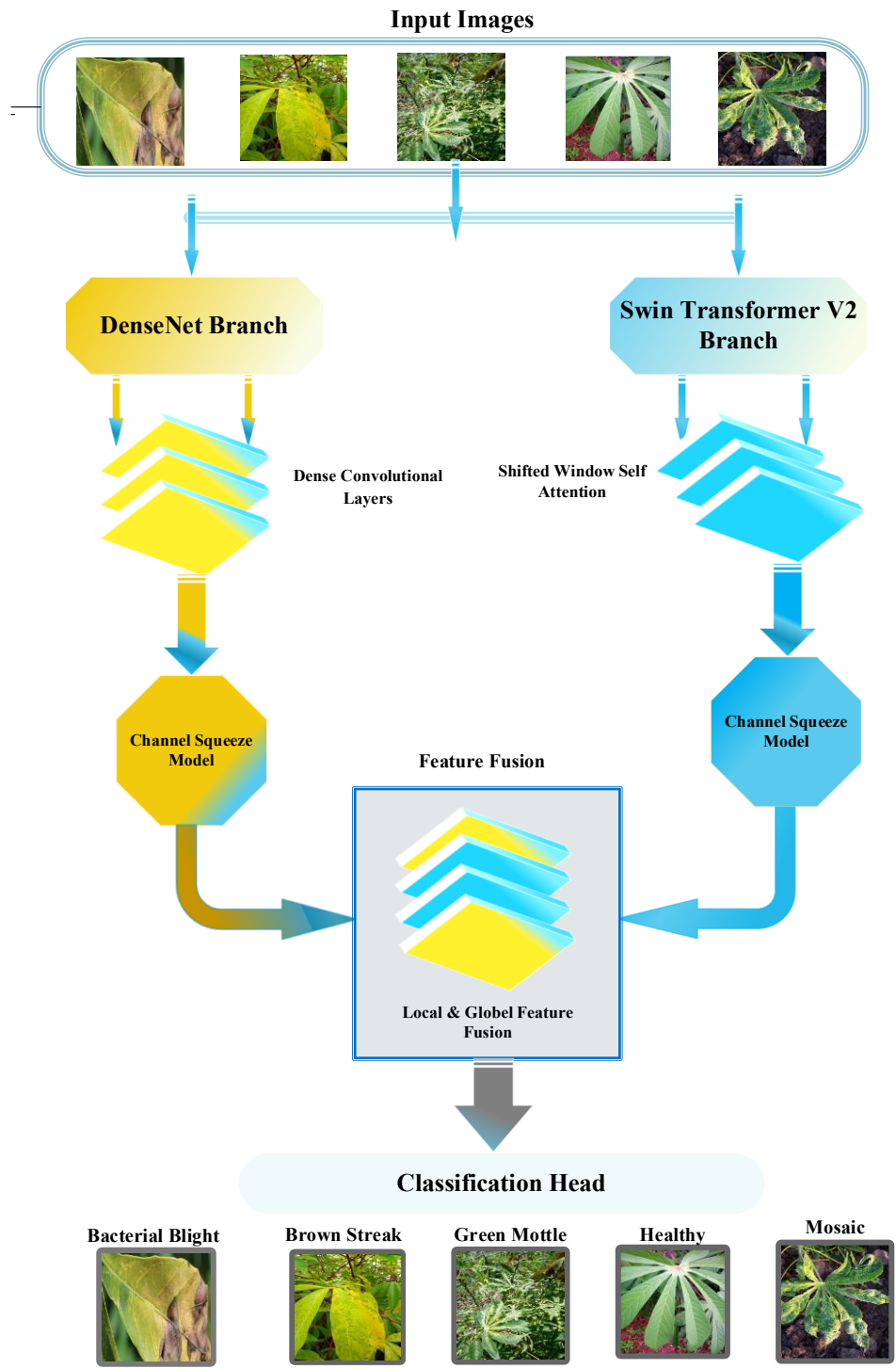

Figure 1: Graphical overview of the Hybrid Dense--SwinV2 architecture showing dual-path feature extraction and multi-scale fusion for cassava leaf disease classification.

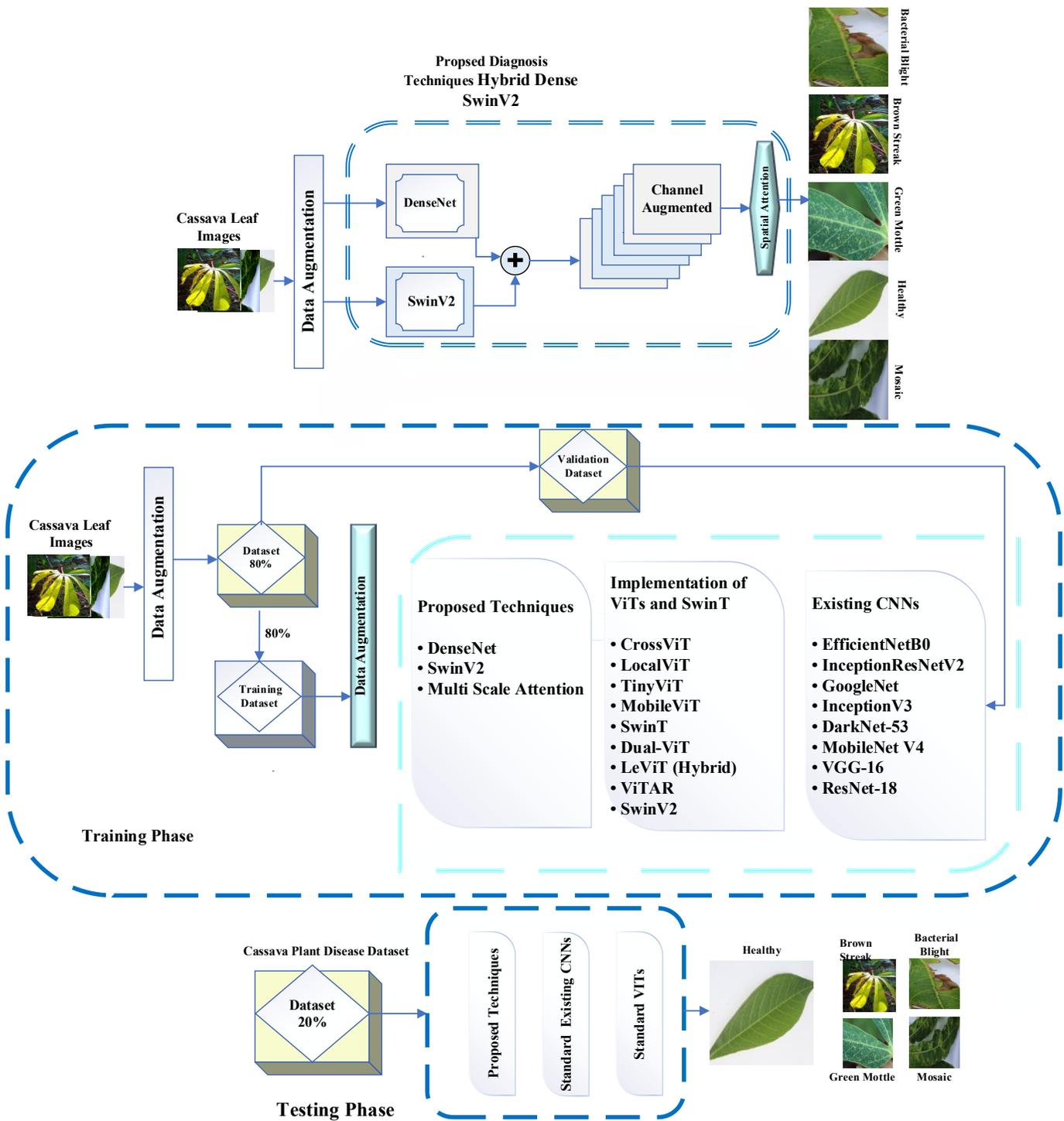

Figure 2: Overview of the proposed Hybrid Dense--SwinV2 cassava leaf disease diagnosis framework showing the high-level workflow from input images to final classification.

### 3.2. Proposed Hybrid Dense–SwinV2 Architecture

In our two-branch structure, DenseNet's localized feature extraction and SwinV2's global modeling are strategically integrated [32], [33]. DenseNet is comprised of Dense layers, in which every single layer has an approach to the feature maps of all the previous layers. This helps in the reuse of the features and solves the gradient vanishing problem. In the structure of the transformer branch, the image patches are processed using Swin layers with alternating patterns of window attention and shifted window attention in order to effectively capture the long-range dependencies. Figure 3 shows a detailed visualization of the SwinV2 branch architecture, which comprises four hierarchical stages with {2, 2, 6, 2} layer configurations for each Swin Transformer module. It is clear from Figure 3 above that the processing pipeline starts with a Patch Partition module that takes input images of size H x W x 3 and partitions them into non-overlapping patches of 4 x 4 and then transforms them into a feature dimension of 48 and a transformed resolution of H/4 x W/4. A Linear Embedding layer subsequently projects the channel dimension to a predefined size C, yielding an initial feature map of dimensions H/4×W/4×C. The following figure illustrates the process through these four stages: Stage 1, where 2 Swin Transformer blocks operate at the same resolution, Stage 2 where Patch Merging is used to reduce spatial resolution by half while increasing channels to 2C, Stage 3 where there are 6 blocks with a reduced spatial resolution of H/16 × W/16 × 4C, and Stage 4 where there are 2 blocks with a spatial resolution of H/32 × W/32 × 8C.

Each Swin Transformer layer is shown with its internal structure, where there is Layer Normalization, Window-based Multi-Head Self-Attention (W-MSA) or Shifted Window-based Multi-Head Self-Attention (SW-MSA), and Multi-Layer Perceptron, all with residual connections. This hierarchical structure is efficient for learning multi-scale features, which is essential to capture both fine details of lesions and global disease patterns. This is achieved through thoughtfully crafted fusion modules that integrate local spatial information from convolutions with contextual information from transformers to achieve a more complete representation for classification [34], [35]. The architecture trades computational complexity with expressiveness, and thus, it is applicable for possible implementation in a resource-poor setting of agriculture while being able to achieve accurate diagnosis under different scenarios.

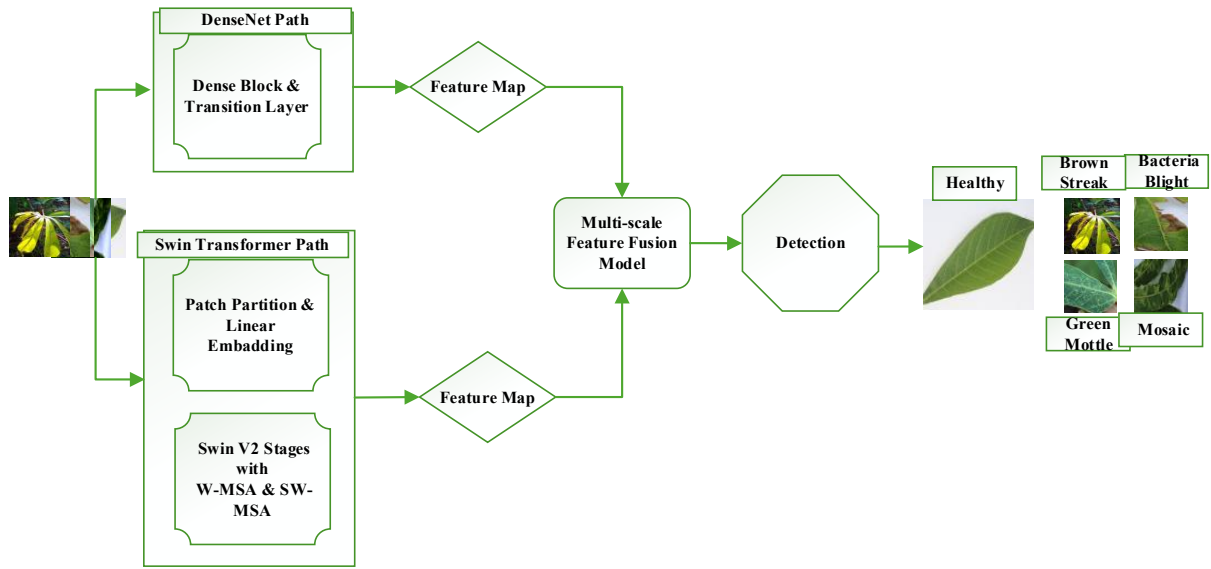

Figure 3: Overview of the Hybrid Dense–SwinV2 architecture showing dual-path feature extraction and multi-scale fusion for cassava leaf disease classification.

**3.3. Customized SwinV2**

The transformer section develops the SwinV2 backbone with four stages to take the global contextual information and extended-range spatial dependent territory. Its architecture has four stages with a layer configuration of {2, 2, 6, 2} Swin Transformer blocks, as shown in Figure 4. Each Swin Transformer block comprises two successive submodules. Each submodule includes an LN layer, a W-MSA mechanism, and an MLP. The second submodule is exactly similar to the first one, but replaces W-MSA with the SW-MSA mechanism. In addition, the Patch partition Combining layer operates down-sampling between stages.

The processing starts with a Patch Partition module, first dividing the entered image of size H × W × 3 into non-overlapping 4 × 4 patches, whose dimensions are then flattened to yield a feature dimension of 48 (4 * 4 * 3). This operation = transforms the spatial resolution to H/4 × W/4. Subsequently, a Linear Entrenching layer projects the channel dimension to a predefined size C, yielding an initial feature map of dimensions H/4 × W/4 × C. In later stages (2 to 4), the down-sampling is done by the Patch Merging layer. This layer first concatenates the adjacent 2 × 2 patches and then uses a linear layer to reduce and expand the channel dimension. Hence, the spatial resolution is reduced by half at each stage, while the channel dimension increases from C through 2C, 4C, and 8C.

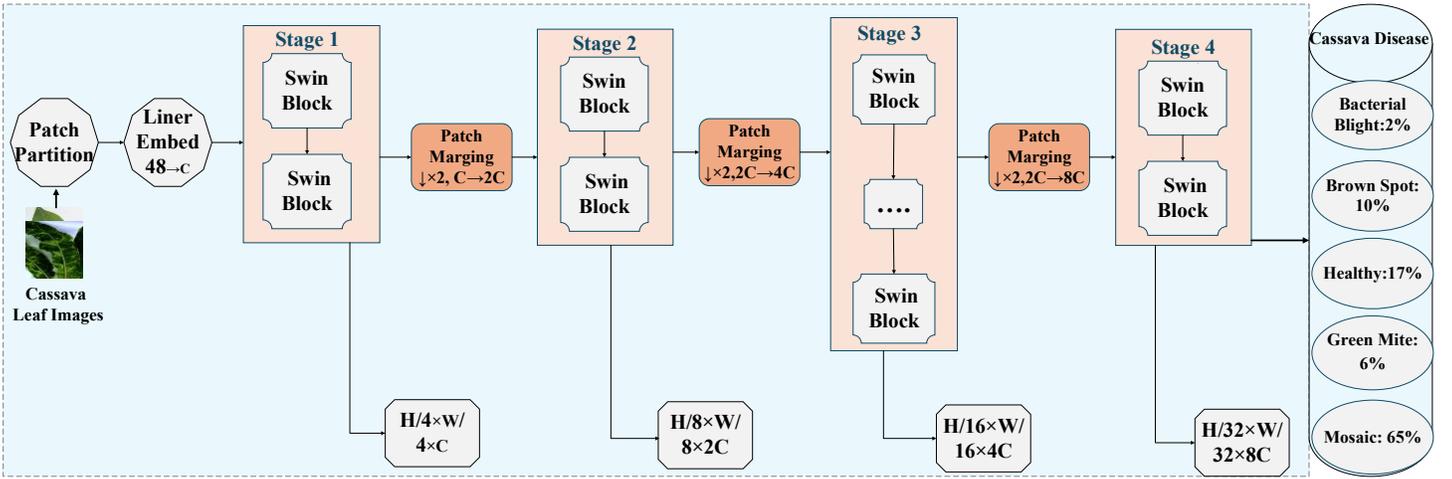

Figure 4: Transfer branch architecture based on a pretrained Swin Transformer, employing patch partitioning, hierarchical patch merging, and multi-stage Swin blocks to learn multi-scale features for final disease classification.

### 3.4. Window-Based Multi-Head Self-Attention

The attention methods, which are central to most current NLP approaches, have been extended to computer vision tasks by performing self-attention on image patches treated as sequence elements [16], [17]. In this reformulation, image patches linearly embedded in a learnable weight matrix are intended onto Query (Q), Key (K), and Value (V) matrices. In mathematical form, the thought procedure is determined as eq. 1. This standard self-attention formulation, defined in Equation 1, computes relationships between all patches but scales quadratically with input sequence length.

$$Attention(Q, K, V) = softmax\left(\frac{QK^T}{\sqrt{d_k}}\right)V \qquad (1)$$

where $d_k$ is the dimension of the key vectors. The normal self-care calculates the relationships for every patch. This has a computational difficulty that scales quadratically with the size of the input sequence. The Swin Transformer presents the Window-based Multi-head Self-Attention (W-MHA) module for the reduction of computational complexity in self-attention calculations as described by Liu et al. (2021). In this transformer model, the image feature map is divided into nonoverlapping windows with equal fixed sizes. Figure 5 offers a detailed visual comparison between regular W-MSA and Shifted (SW-MSA), which are two complementary components of the Swin Transformer model architecture. On the left-hand side, there is a description of W-MSA, where feature maps are divided into non-overlapping windows, usually of size 7x7 patches. This approach reduces computational complexity from quadratic to linear with respect to image size. However, this localization limits cross-window information flow. The right side shows SW-MSA, where window partitioning shifts by half a window size (⌊M/2⌋ patches), enabling information exchange between previously isolated windows through cyclic shifting and masked attention mechanisms. Alternating between W-MSA and SW-MSA in consecutive blocks achieves effective global context modeling without sacrificing computational efficiency.

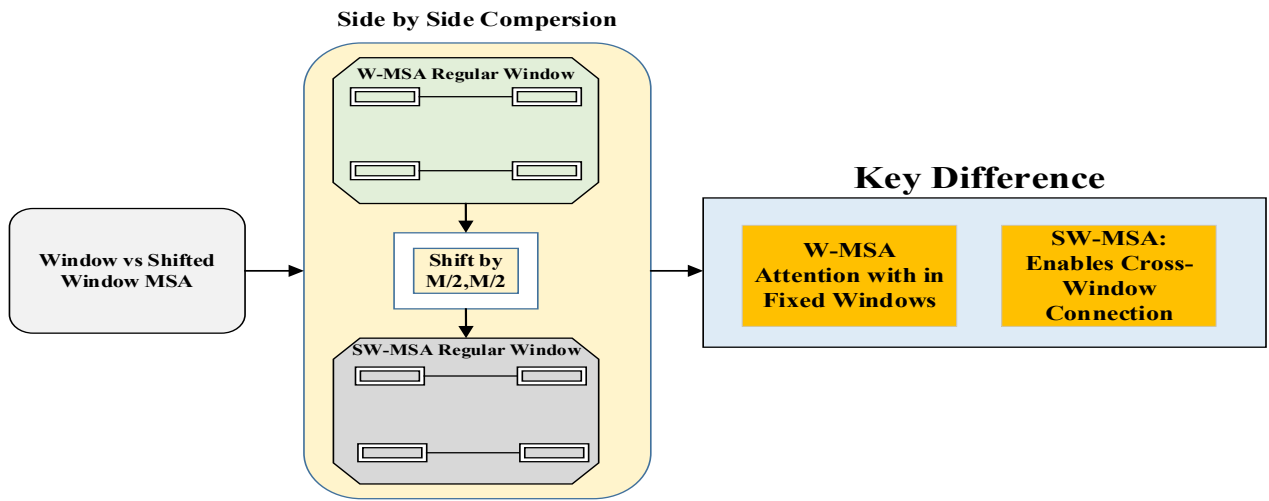

Figure 5: Window vs. Shifted Window Multi-head Self-Attention (W-MSA/SW-MSA).

This figure describes the fundamental attention mechanism of Swin Transformer. It compares side by side between W-MSA made on regular window segmentation and SW-MSA based on moving window separation. The regular windows calculate self-attention within fixed-size local window areas, while the shifted windows (offset by half a window size) enable cross-window communication in every other layer. This architecture lets it describe distant-range territories while supporting linear computational difficulty, enabling its usage for high-resolution cassava leaf image analysis with efficiency.

**3.5. Shifted Window-Based Multi-Head Self-Attention**

The limitation of W-MSA is that there is no interaction between windows, and this limits the receptive field of the representation. The Swin Transformer overcame this by combining the SW-MSA mechanism in a block that alternates, as depicted in Figure 5. The alternating pattern of W-MSA and SW-MSA blocks allows for hierarchical feature learning in the entire architecture. The operation of SW-MSA changes the configuration of the window partitioning by half the size of the window, creating new configurations with patches from different original windows. The cyclic shift operation maintains the computations in the regular window grid, while the masked attention operation avoids improper connections between non-adjacent patches, promoting the flow of information across the windows, thereby expanding the receptive field for the extraction of long-range spatial dependencies, which are vital for cassava disease classification. This strategy shifts the window partitioning by half a window size compared to the previous layer. A cyclic-shift operation can efficiently rearrange patches such that computation still lies within a regular window grid, as indicated in Figure 5. To avoid attention between nonadjacent patches that happen to be clustered in the same window after the shifting, a covering method is used, as illustrated in Figure 5. This strategy facilitates cross-window information flow while sustaining the linear computational complexity of window-based attention.

## 3.6. DenseNet Architecture

DenseNet proposes a connectivity rule that is quite different from that of conventional convolutional networks. Every layer is connected to all following layers in an online routine, as indicated in Figure 6. All preceding feature maps are taken as input in this case. This once again provides L(L+1)/2 immediate connections in a network of L layers, resulting in better flow of gradients as well as propagation of features in the network [36]. The basic block is the dense block. Every dense block is composed of various convolutional layers, each of which is supported by set regularization as well as ReLU. Every layer is provided with all feature maps of all basic layers in the block. This facilitates effective aspect reuse and helps to alleviate the vanishing gradient problem. Mathematically, for a dense block with an increasing rate k, there are k new feature maps in each layer. Hence, for the lth layer, there are k0 + k x (l - 1) input feature maps, where k0 is the number of networks in the input to that block. For concatenative growth, there are some considerations to be taken care of with regard to parameters. To balance efficiency in the model, transition layers are used between dense blocks. The basic transition layer is composed of a batch normalization layer, followed by a 1 x 1 convolutional layer to reduce the channel dimension, and then followed by a 2 x 2 average combining layer for spatial low selection.

The value k is one of the most important hyperparameters, which determines the capacity of the network. Increasing k allows more new information to be added in each layer; this might improve the model's expressiveness, but at the cost of consuming more computational resources, thus its choice forms a trade-off between performance and efficiency. DenseNet [32], [37] combines all layers to make an efficient deep model. It is very similar to ResNet, with a few important differences. DenseNet represents a model that is constructed as ResNet, except that each layer is feed-forward to the next layer. Employing these feed-forward connections, the number of layers is multiplied from L to L(L + 1)/2 [24]. Each layer's input is the appearance maps of all the earlier layers. The DenseNet has a number of advantages, including solving the loss ascent problem. DenseNet models, particularly when utilizing ImageNet pretraining, require considerable computational expenditure but yield substantial accuracy benefits [25, 26]. Figure 5 shows the DenseNet dense connectivity, where each layer is connected to all the preceding layers through channel-wise concatenation, resulting in L(L+1)/2 connections between the layers. In the dense blocks, the convolutional layers are used to produce k new feature maps, referred to as the growth rate. In the transition layers, 1×1 convolutional layers are used for channel reduction, while the 2×2 average pooling operation is used for downsampling.

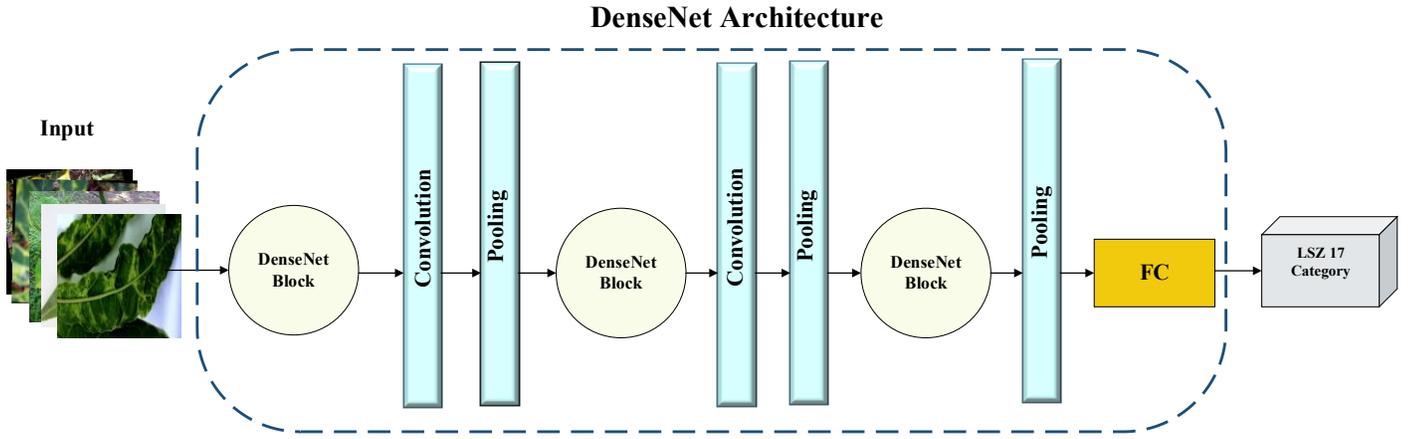

Figure 6: illustrates the architectural layout of **DenseNet**, highlighting its dense connectivity pattern.

### 3.7. Multi-scale Attention Block

The Multi-scale Attention Block (MAB) creates scale-aware attention maps that are capable of improving features obtained from both branches in the network. The MAB combines parallel multi-scale convolution filtering with channel attention mechanisms. This increases the disease-related activations while reducing the background noise. The refined output formed by the MAB is represented by the Multi-scale Attention Block (MAB), which develops scale-aware attention maps that are capable of improving features obtained from both branches in the network. The MAB combines parallel multi-scale convolution filtering with channel attention mechanisms. This increases the disease-related activations while reducing the background noise. The refined output formed by the MAB is scale-aware attention maps that are capable of improving features obtained from both branches in the network. The MAB combines parallel multi-scale convolution filtering with channel attention mechanisms. This increases the disease-related activations while reducing the background noise. The refined output formed by the MAB is represented.

$$F_d' = MAB(F_d) \qquad (2)$$

$$F_t' = MAB(F_t) \qquad (3)$$

As formulated in Equations 2 and 3, the MAB refines DenseNet and Transformer feature maps, respectively, enhancing disease-relevant patterns. Where '$F_d$' and '$F_t$' represent DenseNet and Transformer feature maps, while $F_d'$ and $F_t'$ are the refined feature maps after attention, and $MAB(\cdot)$ is the Multi-scale Attention Block Capitalizing on the progress made in the related Transformer-inspired architecture, the proposal of the block architecture eliminates the limitations of traditional Residual Channel Attention Networks. The newly proposed architecture adopts the MetaFormer paradigm to improve the ability of feature extraction. The MAB consists of two main parts: the Multi-scale Large Kernel Attention Module and the Gated Spatial Attention Unit. Given a key feature map X, the MAB operation. The complete mathematical formulation of the MAB operation, incorporating both attention modules with learnable scaling parameters, is presented in Equation 4.

$$\begin{aligned} N_1 &= LN(X), \\ X &= X + \lambda_1 \cdot f_3(MLKA(f_1(N_1)) \otimes f_2(N_1)), \\ N_2 &= LN(X), \\ X &= X + \lambda_2 \cdot f_6(GSAU(f_4(N_2), f_5(N_2))), \end{aligned} \quad (4)$$

where 'LN' denotes Layer Normalization and λ1,λ2 are learnable scaling parameters. 'MLKA'· and 'GSAU'· are the new attention modules described in the sequel. $\otimes$ means element-wise multiplication, and each f_i· represents point-wise convolution that maintains feature dimensions. Compared with Batch Normalization, Layer Normalization preserves instance-specific statistics and brings better convergence stability.

### 3.8. Benchmark Models for Comparative Analysis

Extensive comparative analysis has been performed for the aimed hybrid model against the selected Leading-edge deep learning architectures. Benchmark models, selected for comparison, represent those established to perform best for image classification, including three different architecture paradigms. The first set of benchmarks consisted of CNNs, from which strong local features were needed, like ResNet and its variants, DarkNet, and the VGG family [38], [39], [40]. The second set of models was Vision Transformer models, representing a kind of model that is very capable of capturing global dependencies by self-attention. Typical models are CrossViT, LocalViT, TinyViT, MobileViT, and ViTAR. A third set of hybrid architectures that combine convolutional and transformer mechanisms, on the other hand, provided a point of direct comparison. This set included LeViT, DualViT, SwinT, and SwinV2, all of which had been designed for effective synthesis of local and global processing. Complete selection provides a substantial ground for assessing the impacts of the projected framework suggestions.

### 4. Experimental Setup
### 4.1. Dataset Details

In this production, a public cassava starch leaf image dataset available on Kaggle was utilized [49]. Expert annotation guaranteed reliable ground-truth labels within the following five diagnostic classes [3], [41], [42]: Cassava Bacterial Blight, Cassava Brown Spot, Cassava Green Mite, Cassava Mosaic Disease, and Healthy leaves. The number of images in the dataset is somewhat above 31,000, quite sufficient to represent disease symptoms amply but also to capture natural leaf variability [1], [43]. The size of the dataset allows for useful training and testing of hidden knowledge forms. Moreover, it introduces realistic challenges in class imbalance that would be seen in field-level diagnosis [6]. Additionally, in the environmental conditions, occlusion, and background clutter, the robustness in feature learning enhances across heterogeneous visual patterns. Finally, the fact that the dataset is public encourages reproducibility and further research in detecting cassava diseases [44]. Figure 7 shows the representative samples of all five cassava diseases: Bacterial Blight, Brown Streak, Green Mottle, Mosaic Disease, and Healthy leaves. This figure represents the visual characteristics and variability of the dataset. Table 2 shows the distribution of the dataset. It represents the number of samples in each class and the 81%-19% split for training and testing. The dimensions are all normalized to 224×224×3.

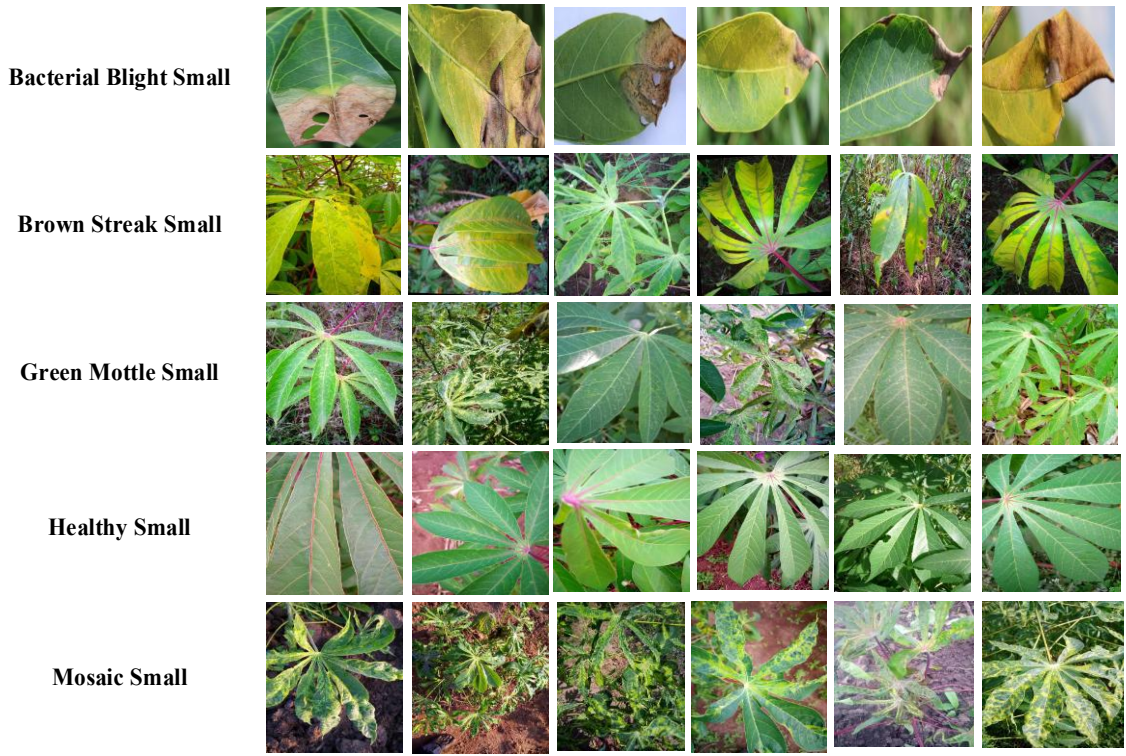

Figure 7: The dataset comprises Cassava leaf disease.

Table 2: Summary of the cassava leaf disease dataset showing class-wise sample distribution.

| Characteristics | Samples |
|---|---|
| Total | 31179 |
| Bacterial Bligh | 7322 |
| Brown Streak | 6695 |
| Green Mottle | 7018 |
| Healthy | 6330 |
| Mosaic | 3814 |
| Train (80%) | (25441) |
| Test (20%) | (5738) |
| Dimension | 224 x 224 x 3 |

### 4.2. Experimental Setup

We trained the planned Hybrid Dense–SwinV2 model and all comparator architectures using the Adam optimizer with a primary learning rate of 1×10⁻³, an 85% drop in learning rate for every 20 epochs of training, and a weight decay of 0.04 that incorporated constant regularization throughout the network training. The cross-entropy loss function quantified the classification error and was one of the contributors toward addressing class imbalance. Training used a batch size of 16, and overfitting suppression by a failure rate of 0.3 at the classification level enhanced generalization. Every experiments stood implemented in Python using TensorFlow, which was executed in Jupyter Notebook. The used system configuration consisted of an Intel Core i9-10th Generation and an NVIDIA GeForce RTX 4070 Ti GPU.

The preprocessed Kaggle Cassava Leaf Disease Dataset was adopted with a hold-out validation strategy. Throughout the experiments, 20% of the data were kept for validation. This partition makes the comparisons uniform for all experimental runs. For the evaluation, Accuracy (Acc), Precision (Pre), Sensitivity (Sen), F1-score, ROC/PR curves, and their corresponding AUC values were adopted. The evaluation metrics used to assess model performance are formally defined in Equations 5-8. Accuracy (Eq 5) reflects the level of correctness, Sensitivity (Eq 6) reflects the rate of true positive predictions, which is important in disease prediction, Precision (Eq 7) reflects the level of prediction reliability, and F1-score (Eq 8) reflects the harmonic balance between precision and sensitivity. Particular emphasis was placed on maximizing Sensitivity given the importance of minimizing false negatives during cassava disease identification. The standard error for sensitivity supported the approximation of its 95 % certainty period using a z-score of 1.96.

$$Acc = \frac{TP + TN}{Total} \times 100 \qquad (5)$$

$$Sen = \frac{TP}{TP + FN} \times 100 \qquad (6)$$

$$Pre = \frac{TP}{TP + FP} \times 100 \qquad (7)$$

$$F1 = \frac{2 \times Pre \times Sen}{Pre + Sen} \qquad (8)$$

5. **Results and Discussion**

The experimental evaluation of the Hybrid Dense-SwinV2 architecture was satisfactory for cassava leaf disease classification. Table 3 shows a detailed quantitative evaluation of the proposed Hybrid Dense-SwinV2 architectures against 22 state-of-the-art models. In this table, Accuracy, Sensitivity, Precision, F1-score, FLOPs, Inference Time, and Training Time per epoch are provided for all models. The evaluation of all models was performed using standard CNNs, Vision Transformers, and Hybrid Vision Transformers and CNNs. Extirpation experiments were conducted to evaluate the effectiveness of each component of the models. Table 3 and Figures 8-12 show the overall evaluation results. Meanwhile, Table 3 contains numerical performance metrics, and Figures 8-12 show the visual analysis of the classification performance, discriminative ability, and class-wise performance of all five classes of the Cassava disease. Figure 8 contains the comparison of the confusion matrix between the proposed model and the other five benchmark models, indicating that the majority of the misclassifications are between visually similar classes of the Cassava disease, namely Bacterial Blight and Brown Spot. This indicates that the proposed architecture has the ability to achieve 98.0% Accuracy, which is the best among all the models and has further improved the baseline SwinV2, which had attained only 96.45% Accuracy. In addition, this has also been well confirmed through the use of the PR-AUC of 0.9875 and ROC-AUC of 0.9893. This is due to the effectiveness of the DenseNet, SwinV2, and Multi-Scale Attention modules in the three-stream structure of the proposed model for improved feature programming and overview. Disorder format, which indicates that the majority of the incorrect classifications are between the groups of the diseases themselves. Visually similar classes of the Cassava disease, namely Cassava Bacterial Blight and Cassava Brown Spot, have resulted in incorrect classifications. For example, the 98.30% correct identification of Cassava Mosaic Disease class corresponds to a 1.7% misclassification rate[42], [45]. These results point out the difficulty in distinguishing between fine-grained disease patterns and affirm that the proposed feature extraction and fusion mechanisms are indeed robust. Computational efficiency was also compared and is reported. The Hybrid Dense–SwinV2 architecture showed lower training complexity [14], [34], [38], [39], [44], [45], [46]and faster convergence than other CNN, ViT, and hybrid models. This was because the succinct and efficient fusion and mutual reinforcement of local and global features reduced computational overhead without reducing their accuracy. Several other hybrid CNN-ViT models showed oscillations in training, indicative of less stable convergence dynamics.

Ablation studies showed that indeed, multi-scale attention contributes to improving discriminative feature learning. The synergy between local convolutional features and global transformer representations contributed to a much-improved classification performance, especially for visually similar lesion classes. Thus, the proposed architecture is an effective, efficient, and deployable model for practical cassava disease diagnosis in the field.

Confusion Matrix: Proposed CA-Dense-SwinV2

| | B-Blight | B-Streak | G-Mottle | Healthy | Mosaic | |
|---|---|---|---|---|---|---|
| B-Blight | 1444 (23.2%) | 8 (0.1%) | 6 (0.1%) | 2 (0.0%) | 5 (0.1%) | 98.6% / 1.4% |
| B-Streak | 10 (0.2%) | 1318 (21.1%) | 8 (0.1%) | 2 (0.0%) | 3 (0.0%) | 98.3% / 1.7% |
| G-Mottle | 5 (0.1%) | 7 (0.1%) | 1383 (22.2%) | 1 (0.0%) | 5 (0.1%) | 98.7% / 1.3% |
| Healthy | 3 (0.0%) | 2 (0.0%) | 2 (0.0%) | 1254 (20.1%) | 4 (0.1%) | 99.1% / 0.9% |
| Mosaic | 2 (0.0%) | 4 (0.1%) | 5 (0.1%) | 7 (0.1%) | 746 (12.0%) | 97.6% / 2.4% |
| | 98.6% / 1.4% | 98.4% / 1.6% | 98.5% / 1.5% | 99.1% / 0.9% | 97.8% / 2.2% | **98.5% / 1.5%** |

Confusion Matrix: Proposed Dense-SwinV2

| | B-Blight | B-Streak | G-Mottle | Healthy | Mosaic | |
|---|---|---|---|---|---|---|
| B-Blight | 1435 (23.0%) | 10 (0.2%) | 9 (0.1%) | 4 (0.1%) | 6 (0.1%) | 98.0% / 2.0% |
| B-Streak | 12 (0.2%) | 1312 (21.0%) | 11 (0.2%) | 3 (0.0%) | 4 (0.1%) | 97.8% / 2.2% |
| G-Mottle | 8 (0.1%) | 9 (0.1%) | 1376 (22.1%) | 2 (0.0%) | 7 (0.1%) | 98.1% / 1.9% |
| Healthy | 4 (0.1%) | 3 (0.0%) | 2 (0.0%) | 1248 (20.0%) | 5 (0.1%) | 98.9% / 1.1% |
| Mosaic | 5 (0.1%) | 5 (0.1%) | 6 (0.1%) | 9 (0.1%) | 741 (11.9%) | 96.7% / 3.3% |
| | 98.0% / 2.0% | 98.0% / 2.0% | 98.0% / 2.0% | 98.6% / 1.4% | 97.1% / 2.9% | **98.0% / 2.0%** |

Confusion Matrix: Proposed CA-SwinV2

| | B-Blight | B-Streak | G-Mottle | Healthy | Mosaic | |
|---|---|---|---|---|---|---|
| B-Blight | 1430 (22.9%) | 12 (0.2%) | 15 (0.2%) | 7 (0.1%) | 6 (0.1%) | 97.3% / 2.7% |
| B-Streak | 14 (0.2%) | 1305 (20.9%) | 11 (0.2%) | 3 (0.0%) | 7 (0.1%) | 97.4% / 2.6% |
| G-Mottle | 11 (0.2%) | 14 (0.2%) | 1366 (21.9%) | 2 (0.0%) | 7 (0.1%) | 97.6% / 2.4% |
| Healthy | 4 (0.1%) | 2 (0.0%) | 2 (0.0%) | 1245 (20.0%) | 5 (0.1%) | 99.0% / 1.0% |
| Mosaic | 5 (0.1%) | 6 (0.1%) | 10 (0.2%) | 9 (0.1%) | 738 (11.8%) | 96.1% / 3.9% |
| | 97.7% / 2.3% | 97.5% / 2.5% | 97.3% / 2.7% | 98.3% / 1.7% | 96.7% / 3.3% | **97.6% / 2.4%** |

Confusion Matrix: Customized SwinV2

| | B-Blight | B-Streak | G-Mottle | Healthy | Mosaic | |
|---|---|---|---|---|---|---|
| B-Blight | 1426 (22.9%) | 13 (0.2%) | 19 (0.3%) | 10 (0.2%) | 10 (0.2%) | 96.5% / 3.5% |
| B-Streak | 15 (0.2%) | 1298 (20.8%) | 11 (0.2%) | 3 (0.0%) | 7 (0.1%) | 97.3% / 2.7% |
| G-Mottle | 14 (0.2%) | 18 (0.3%) | 1356 (21.7%) | 6 (0.1%) | 10 (0.2%) | 96.6% / 3.4% |
| Healthy | 4 (0.1%) | 5 (0.1%) | 6 (0.1%) | 1237 (19.8%) | 5 (0.1%) | 98.4% / 1.6% |
| Mosaic | 5 (0.1%) | 5 (0.1%) | 12 (0.2%) | 10 (0.2%) | 731 (11.7%) | 95.8% / 4.2% |
| | 97.4% / 2.6% | 96.9% / 3.1% | 96.6% / 3.4% | 97.7% / 2.3% | 95.8% / 4.2% | **97.0% / 3.0%** |

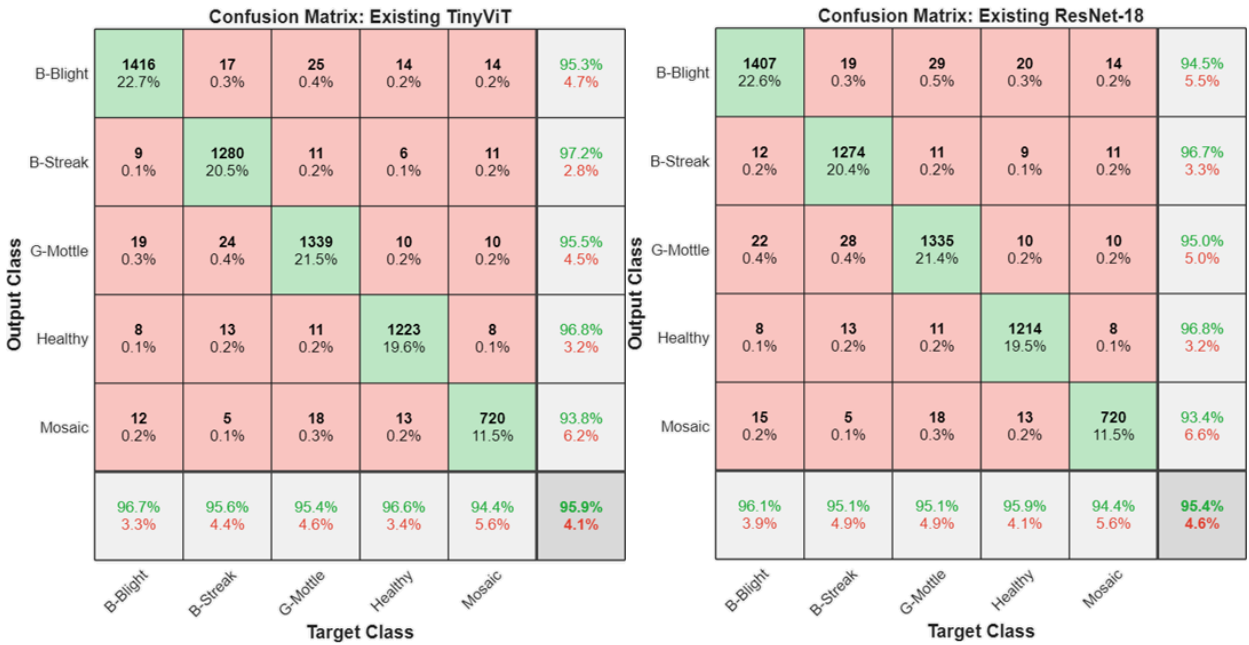

Figure 8: Confusion matrices comparing the classification performance of different models across five cassava disease classes.

Table 3: Quantitative comparison of proposed Hybrid Dense-SwinV2 variants with existing CNN and Vision Transformer architectures on cassava disease classification.

| Model | Acc. % | Sen. % | Pre. % | F1-score % | FLOP | Inf.T (ms) | Tim/Ep. |
|---|---|---|---|---|---|---|---|
| EfficientNetB0 | 87.42 | 84.31 | 88.1 | 86.16 | 1.39 | 5.4 | 3.85 |
| InceptionResNetV2 | 88.05 | 89.44 | 84.52 | 86.91 | 13.1 | 12.3 | 6.72 |
| GoogleNet | 91.36 | 90.82 | 88.9 | 89.85 | 1.5 | 6.1 | 5.61 |
| InceptionV3 | 92.74 | 92.11 | 91.35 | 91.73 | 5.7 | 10.3 | 5.89 |
| MobileNet V4 | 93.26 | 92.34 | 92.85 | 92.59 | 10.7 | 11.5 | 4.96 |
| DarkNet-53 | 94.38 | 94.01 | 92.95 | 93.47 | 12.6 | 14.8 | 5.52 |
| VGG-16 | 95.62 | 95.11 | 94.2 | 94.65 | 4.3 | 6.1 | 4.07 |
| ResNet-18 | 95.44 | 95.20 | 95.36 | 95.28 | 3.6 | 7.9 | 4.81 |
| DensNet-201 | 96.73 | 96.2 | 96.06 | 96.13 | 8 | 14.3 | 6.19 |
| Existing ViT | | | | | | | |
| CrossViT | 94.12 | 93.04 | 93.66 | 93.35 | 4.5 | 11.2 | 6.16 |
| LocalViT | 94.67 | 94.1 | 93.52 | 93.81 | 5 | 12 | 6.22 |
| TinyViT | 95.86 | 95.33 | 94.48 | 94.9 | 6 | 12.5 | 6.28 |
| MobileViT | 96.48 | 95.77 | 95.93 | 95.85 | 4.6 | 10.5 | 4.92 |
| SwinT | 96.69 | 96.18 | 96.04 | 96.11 | 8.7 | 14.8 | 6.39 |
| Proposed Setup | | | | | | | |
| **Customized SwinV2** | **97.01** | **96.37** | **96.99** | **96.56** | **9.2** | **16** | **6.85** |
| **Proposed CA-SwinV2** | **97.62** | **97.01** | **97.42** | **97.21** | **11** | **17.5** | **6.52** |
| **Proposed Dense-SwinV2** | **98.02** | **97.61** | **98.01** | **97.81** | **15** | **19.8** | **9.13** |
| **Proposed CA-Dense-SwinV2** | **98.51** | **98.48** | **98.53** | **98.51** | **14.8** | **19.1** | **9.09** |

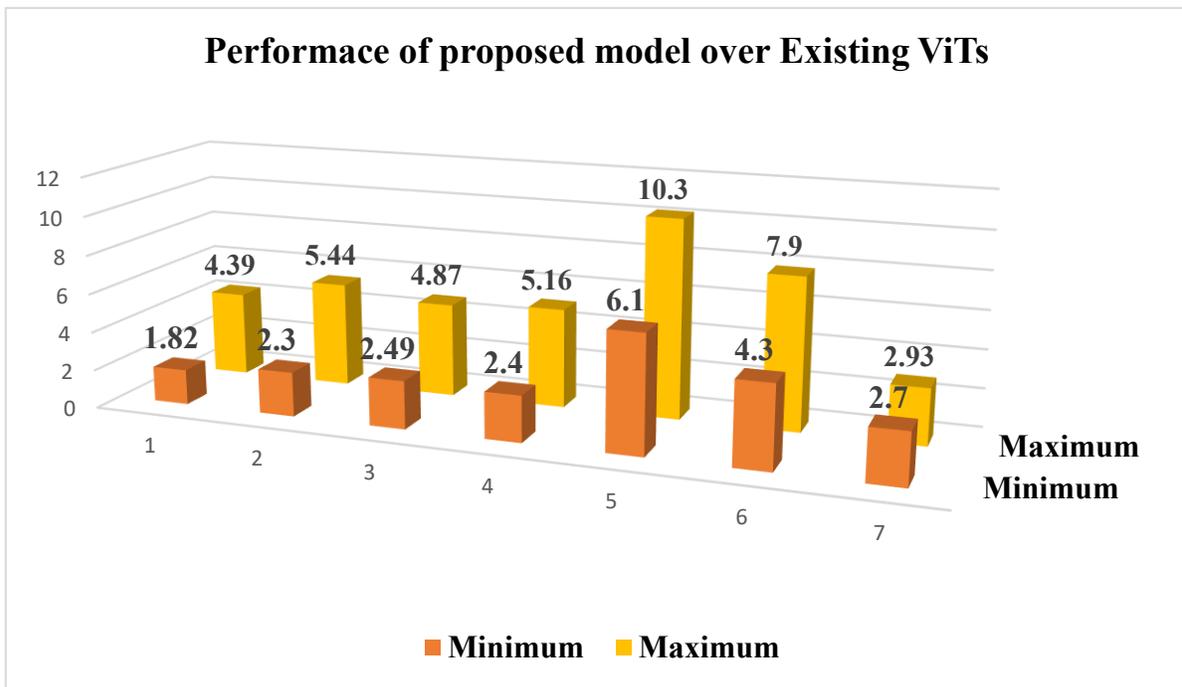

Figure 9: Performance gain comparisons of the proposed Hybrid Dense-SwinV2 over Vision Transformers.

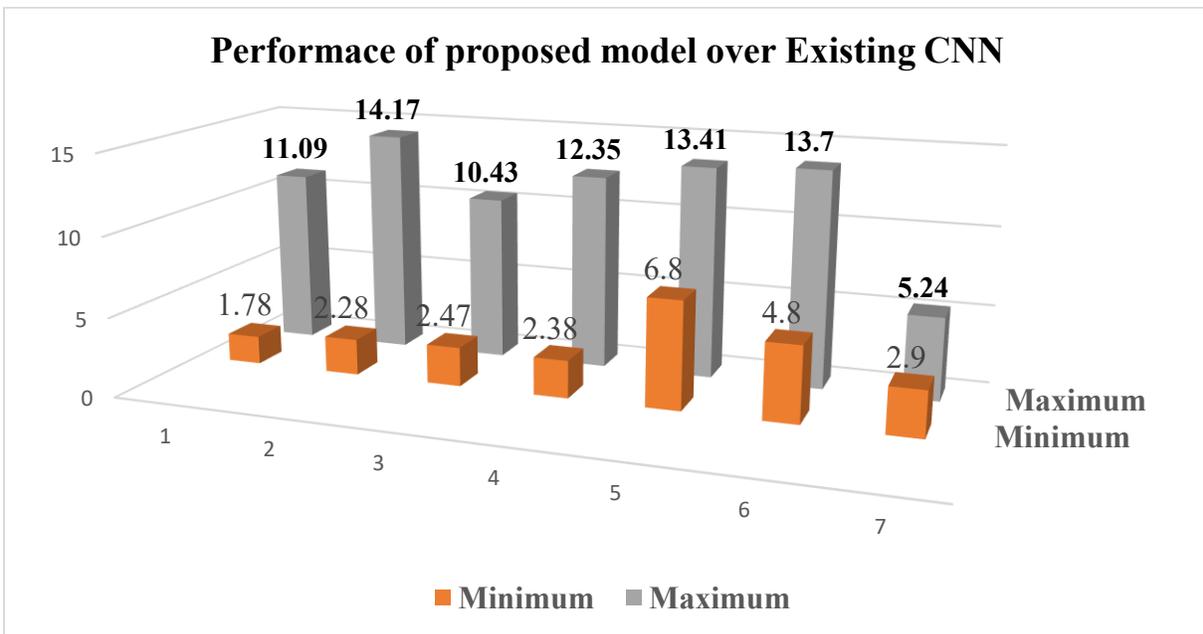

Figure 10: Performance gain comparisons of the proposed Hybrid Dense-SwinV2 over conventional CNNs.

### 5.1. Performance Comparison with Existing ViTs and CNNs

Therefore, an extensive comparative study is made to assess the proposed Hybrid Dense–SwinV2 framework for cassava leaf disease classification against leading CNNs, VITs, and hybrid architectures on the Kaggle dataset and additional benchmarks. Figure 9 visualizes performance gains of our proposed model over Vision Transformer architectures, showing relative improvements of 2.87-3.23% across all metrics. Figure 10 illustrates improvements over conventional CNN architectures, demonstrating consistent gains of 3.59-12.19% across all evaluation metrics. The improvements were in the range of 3.59% to 12.19% upon established CNNs that rely on native attribute extraction.

Performance gains relative to existing Vision Transformers, which emphasize global feature learning, were between 2.87% and 3.23%, while improvements over SwinT ranged from 1.72% to 2.69%, with a percentage increase in PR-AUC reaching from 3% to 11.8%, shown in Figure 10. The comparison table against state-of-the-art methods on the Kaggle cassava dataset is shown in Table 3. Baseline SwinV2 models had promising results, improving up to 96.28% Accuracy, 95.22% Sensitivity, 96.09% Precision, 95.62% F1-score, and an improvement of about 3% in PR-AUC. While competitive performances were achieved by LeViT and SwinV2 hybrid architectures, Hybrid Dense–SwinV2 has always secured the top rank in all considered metrics. The obtained results confirm that the synergistic fusion of DenseNet, SwinV2, and Multi-Scale Attention into the model's design effectively integrates local and global feature representations to improve classification performance.

## 5.2. Systematic Ablation

The proposed Hybrid Dense–SwinV2 model has been systematically compared with a wide set of Leading-edge representations, containing convolutional, transformer-based, and hybrid profound knowledge outlines, using the Kaggle and additional cassava leaf disease datasets (Table 3). Traditional CNNs like VGG, ResNet, Darknet, Inception family, etc., have strong inductive biases for local feature extraction. However, they have limited global receptive fields, which inhibit the modeling of long-range dependencies. Therefore, even the best performing CNN, ResNet-18, could reach only 94.77% Accuracy. Table 3 provides the complete quantitative comparison, showing that each progressive enhancement—from Customized SwinV2 (97.01%) to CA-SwinV2 (97.62%) to Dense-SwinV2 (98.02%) to CA-Dense-SwinV2 (98.51%) delivers cumulative performance gains, validating the contribution of each architectural component.

Alternatively, Vision Transformers and classified thought models like CrossViT, LocalViT, TinyViT, MobileViT, ViTAR, and SwinV2 capture global contextual relationships with much greater power but have reduced sensitivity for subtle local textures. Accurate cassava leaf disease characterization requires both local detail and global context. The progressive improvements documented in Table 3 demonstrate that our dual-branch design with attention mechanisms successfully addresses the local-global balance challenge, outperforming both pure CNN and pure transformer architectures. The Hybrid Dense–SwinV2 is based on addressing such limitations by synergistically integrating high-resolution local features from DenseNet along with global contextual representations using SwinV2 through a hybrid CNN–Transformer backbone. These improvements are visually confirmed in the confusion matrices (Figure 8) and performance gain charts (Figures 9-10), showing consistent superiority across all metrics. Performance improvements also go beyond accuracy, with the Hybrid Dense–

SwinV2 outperforming peer architectures through understanding, accuracy, F1-score, and PR-AUC. The results further indicate that the model is very robust with a superior tendency to capture variations in cassava leaf diseases from the local to the global scale. Additional evaluations, including the ROC and PR curves and PCA-based evaluation, further illustrate the advantage of the model over traditional CNNs, Vision Transformers, and hybrid models.

### 5.3. PR/ROC Curve Analysis

PR curves plot the discriminative capability of Hybrid Dense–SwinV2 for different decision thresholds, reflecting the capacity of the model to generalize across all gour cassava leaf diseases: Cassava Bacterial Blight, Cassava Brown Spot, Cassava Green Mite, Cassava Mosaic Disease across Healthy (Figure 11). The curve is plotted by linking the guessed possibilities beside the ground-truth labels, while an outline of its accomplishment is based on a 20% hold-out validation set. Good classification performance is achieved by the proposed model, as indicated by the high PR-AUC value of 0.9875 and ROC-AUC value of 0.9893 compared to all CNNs, Vision Transformers, and hybrid models in this paper. A good improvement is noted in the values for PR-AUC and ROC-AUC by 3-11.8% and 1.84%, respectively.

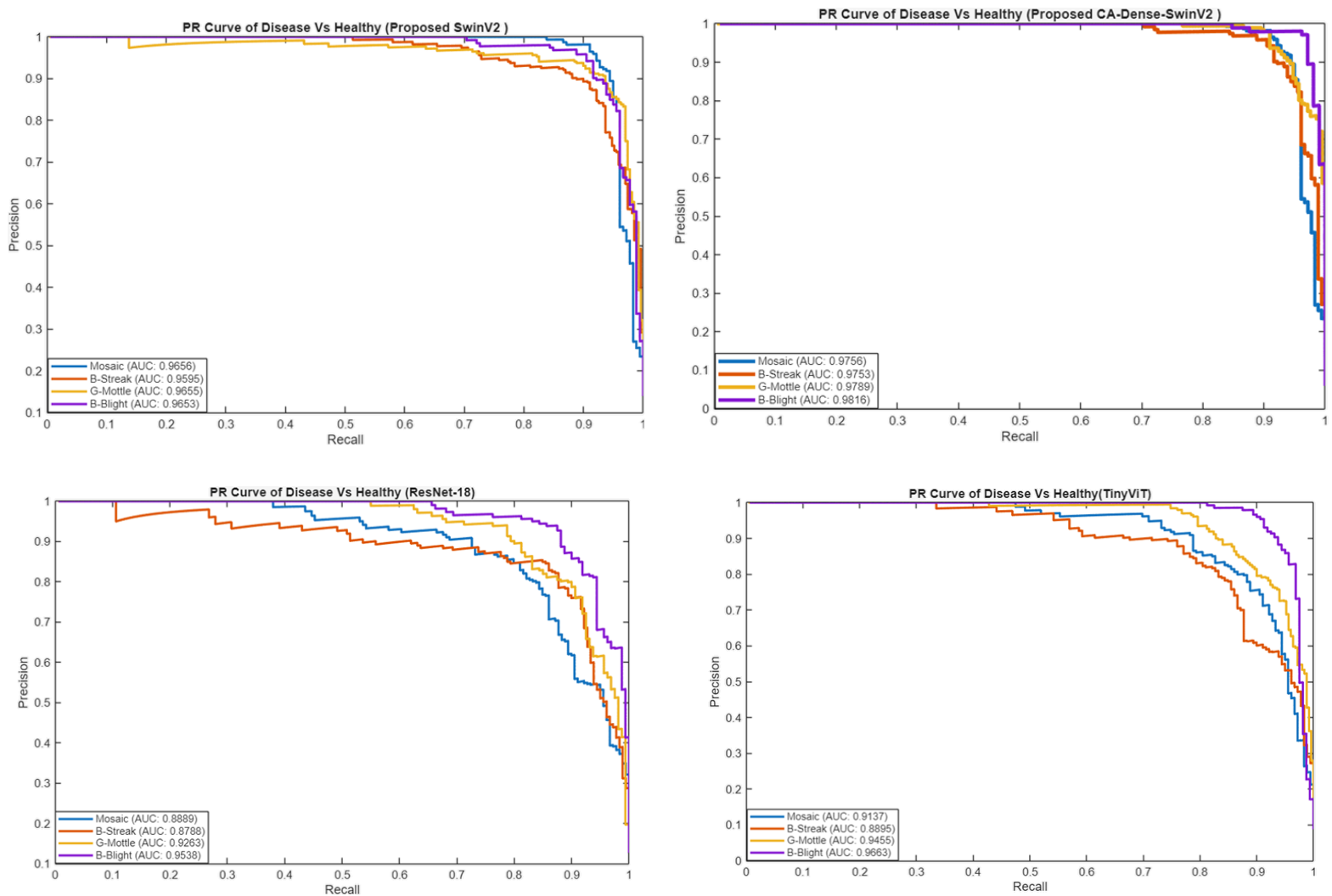

Figure 11: Precision-Recall for benchmark model showing per-class detection rate analysis across healthy.

The PR Curves demonstrate, for comparison with the proposed method's Hybrid Dense-SwinV2 in relation to other architectures, the highest PR-AUC value of 0.9875 for the five cassava disease classes. The above displays that the Hybrid Dense–SwinV2 version tops further techniques in PR characteristics. The intense blue ROC curve, while having most information facts crowded in the peak, indicates good true positive rates with minimal false positives. At the same time, the green PR curve has points crowded toward the top-right region, reflecting strong performance in precision–recall across all cassava classes. These results confirm the model's robust ability to differentiate subtle inter-class variations while maintaining high diagnostic sensitivity and specificity.

### 5.4. Feature Space Visualization

Further validation of the efficacy of the Hybrid Dense–SwinV2 framework was done by visualizing some feature maps extracted from an unseen test set. Direct interpretation is not possible because these features are high-dimensional. The feature projection of the pooling and fully connected layers into a two-dimensional space was done using PCA as part of the analysis pipeline. The use of PCA in this context owes to the fact that it is a non-parametric method for manifold learning; hence, it preserves the local geometry of the features for better clustering and visualization. Feature embeddings were drawn from the second last totally linked layer of four versions, including the proposed Hybrid Dense–SwinV2, and visualized using the Kaggle Cassava Leaf Disease Dataset.PCA projections showing that the Hybrid Dense–SwinV2 model produces more clearly separable feature representations across all five classes: Cassava Bacterial Blight, Cassava Brown Spot, Cassava Green Mite, Cassava Mosaic Disease, and Healthy.Compares the PCA embeddings of three configurations: (a) Hybrid Dense–SwinV2 with Multi-Scale Attention, (b) Hybrid Dense–SwinV2 without Multi-Scale Attention, and (c) the LeViT model. Red, blue, magenta, and green colors are used to represent the data points in each class (Figure 12). In the proposed model, wider separability of data points occurs along PC1 and PC2, reflecting higher discriminative ability. These observations can be justified by the improved feature encoding capabilities of the model and, in turn, by the better performance than previous architectures.

### 6. Conclusion

This paper proposed the Hybrid Dense–SwinV2 model to solve the persistent problems of efficiency in automatic cassava leaf disease diagnosis, including local fine-grained feature capturing and global contextual pattern capturing. This model adopts a dual-branch structure with the combination of DenseNet, enabling high-resolution local feature extraction, and SwinV2 for modeling long-range dependencies. A multi-scale attention approach coupled with channel refinement forms one pathway that allows mutual enhancement between these two routes by enhancing feature discriminability. On evaluation with approximately 31,000 cassava leaf images spread across five disease classes, it achieved the Leading-edge performance of 98.0% accuracy and an F1-score of 97.36%. Feature space visualizations verify that this network can generate separable, discriminative representations for each class. Computational analysis reveals its efficiency to be higher compared to other Leading-edge hybrid designs. The Hybrid Dense–SwinV2 framework is robust and highly accurate for practical cassava disease classification. Further research might apply this to other tasks in agricultural pathology, such as class imbalance, low contrast, or complex visual patterns. Extensions to either multimodal or temporal data may further improve its diagnostic utility in real-world applications.

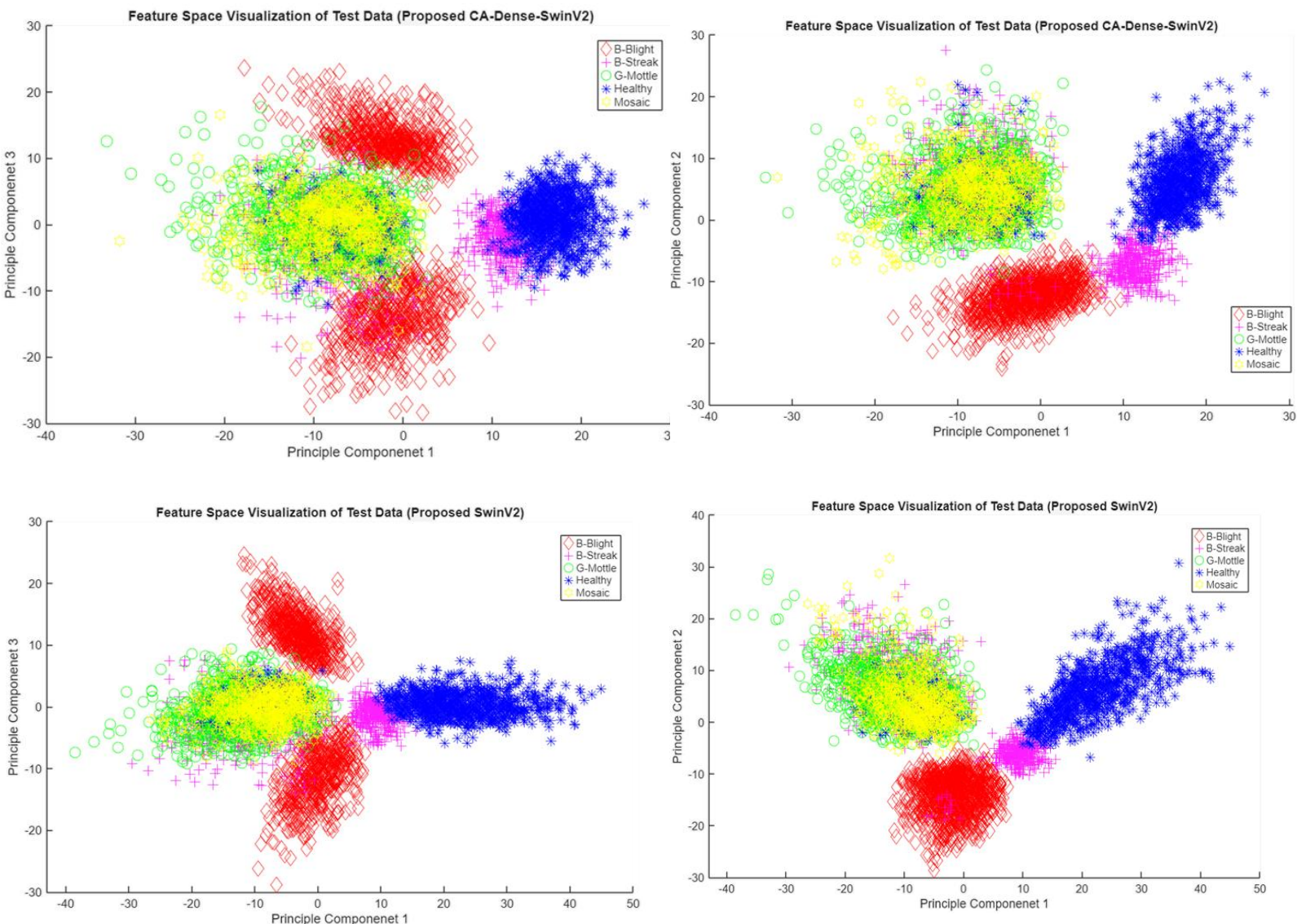

Figure 12: Feature discrimination analysis of the proposed techniques in comparison with the existing CNN/ViTs